\documentclass[runningheads]{llncs}

\usepackage{amssymb}
\usepackage{graphicx}
\usepackage{algorithm}
\usepackage{algorithmic}
\usepackage{subfigure}
\usepackage{figsize}
\usepackage{verbatim} 
\usepackage{multirow}

\usepackage{url}
\urldef{\mailsa}\path|{shai.haim, toby.walsh}@nicta.com.au|

\begin{document}

\mainmatter

\title{Online Estimation of SAT Solving Runtime\thanks{The second author is funded by
DCITA and the ARC through Backing Australia's Ability
and the ICT Centre of Excellence program.}}

\titlerunning{Online Estimation of SAT Solving Runtime}

\author{Shai Haim\and Toby Walsh}

\authorrunning{S. Haim and T. Walsh}

\institute{NICTA and UNSW\\
\mailsa\\}

\maketitle

\begin{abstract}

We present an online method for estimating the cost of solving SAT problems. 
Modern SAT solvers present several challenges to estimate 
search cost including non-chronological backtracking, learning and restarts. 
Our method uses a linear model trained on data gathered
at the start of search. We show the effectiveness of this method 
using random and structured problems. We demonstrate 
that predictions made in early restarts can be used to 
improve later predictions. We also show that we can 
use such cost estimations to select a solver from a portfolio.
\end{abstract}

\section{Introduction}

Modern SAT solvers present several challenges for 
estimating their runtime. For instance, clause learning 
repeatedly changes the problem the solver faces. 
Estimation of the size of the search tree
at any point should take into consideration 
the changes that future learning clauses will cause. 
As a second example, restarting generates a new search tree
which again makes prediction hard. 
Our approach to these problems is to 
use a machine learning based on-line method to predict the cost of the 
search by observing the solver's \emph{behaviour} 
at the start of search.

\label{WBE-RE}

Previous methods include the Weighted Backtrack Estimator,
the Recursive Estimator (\cite{kil:sla:thi:wal}) and 
the SAT Progress Bar (\cite{kok:shl}) that do not support backjumping or restarts, and the 
BDD-based Satometer (\cite{alo:sie:sak}) which doesn't provide an estimate for the size of the decision tree. 
\label{ml-background}
Machine learning has also been used to estimate search cost. 
Horovitz et al  used a Bayesian approach to classify CSP and SAT problems 
according to their runtime \cite{hor:rua:gom:kau:sel:chi}. 
Whilst this work is close to ours, there are 
some significant differences. For example, they used SATz-Rand which does not 
use clause learning.
Xu et. al \cite{xu:hoo:ley} used machine learning to tune
empirical hardness models \cite{ley:nud:sho}. The
only non-static features used were
generated by probes of DPLL and stochastic search. 
Their method gives an estimate
for the distribution of runtimes
and not, as here, an estimate for a specific run.
Finally, an online machine learning method has been
used for QBF solvers \cite{sam:mem}.

\section{Linear model prediction (LMP)}

\label{Sec:LMP}

We predict the size of subtrees to follow from
the subtrees explored in the past.
Given a problem $\cal P$$\in E$, when $E$ 
is an ensemble of problems, we first train the model using 
a subset of problems $\cal T $$\subset E$. For every training 
example $t \in \cal T$, we create a feature vector 
$x_{t}=\left\{{x_{t,1},x_{t,2},\ldots,x_{t,k}}\right\}$. 
We select features by removing those with the smallest 
standardised coefficient until no improvement is observed
based on the standard AIC (Akaike Information Criterion). 
We then search for and eliminate co-linear features in the set. 

Using ridge linear regression, we fit our coefficient vector $w$ 
to create a linear predictor $f_{w}\left(x_{i}\right) =w^{T}x_{i}$. 
We chose ridge regression since it is quick and simple,
and generally yields good results. 
We predict the log of the number of conflicts as runtimes
vary significantly.
Since the feature vector is computed online, we 
do not want it to add significant cost to search. It
therefore only contains features that can be calculated in (amortized) constant time. 
We define the \emph{observation window} to be that part of search 
where data is collected. At the end of the observation window,
the feature vector is computed and the model queried for an estimation.
\begin{table}
	
	\caption{The feature vector used by linear regression to construct prediction models}
	\begin{minipage}{\textwidth}
	\centering
	\label{Table:featurevector}
  \begin{tabular}{ | l | c | c | c | c | c | c |}
  \hline
		\multirow{2}{*}{$Feature$} & \multirow{2}{*}{$init$} &\multicolumn{5}{|c|}{\emph{Observation Window}}\\\cline{3-7}
		 &  & $min$ & $max$ & $mean$ & $SD$ &	$last$\\\hline
		Number of $variables$ ($var$) & $\surd$ &  &  &  &  &  \\
		Number of $clauses$ ($cls$) & $\surd$ &  &  &  &  &  \\
		$cls/var$ & $\surd$ & $\surd$ & $\surd$ & $\surd$ & $\surd$ & $\surd$ \\
		$var/cls$ & $\surd$ & $\surd$ & $\surd$ & $\surd$ & $\surd$ & $\surd$ \\
		Fraction of Binary Clauses & $\surd$ &  &  & $\surd$ & $\surd$ & $\surd$ \\
		Fraction of Ternary Clauses & $\surd$ &  &  & $\surd$ & $\surd$ & $\surd$ \\
		Avg. Clause Size & $\surd$ &  &  & $\surd$ & $\surd$ & $\surd$ \\
		Search Depth (from assignment stack)&  &  & $\surd$ & $\surd$ & $\surd$ &  \\
		Search Depth (in corresponding binary tree)\footnote{see \cite{hai:wal} for further details}&  &  & $\surd$ & $\surd$ & $\surd$ &  \\
		Backjump Size & &  & $\surd$ & $\surd$ & $\surd$ &  \\
		Learnt Clause Size &  & $\surd$ & $\surd$ & $\surd$ & $\surd$ &  \\
		Conflict Clause Size &  & $\surd$ & $\surd$ & $\surd$ & $\surd$ &  \\
		Fraction of assigned vars before backtracking ($abb$) &  & $\surd$ & $\surd$ & $\surd$ & $\surd$ &  \\
		Fraction of assigned vars  after backtracking ($aab$) &  & $\surd$ & $\surd$ & $\surd$ & $\surd$ &  \\
		$aab.mean/abb.mean$  &  & $\surd$ & $\surd$  & $\surd$ & $\surd$ &  \\
		$abb.mean/aab.mean$  &  & $\surd$ & $\surd$  & $\surd$ & $\surd$ &  \\
		$log(WBE)$ &  & $\surd$ & $\surd$ & $\surd$ & $\surd$ & $\surd$ \\
		\hline
  \end{tabular}
  \end{minipage}
  
\end{table}

The feature vector measures both problem structure and search behaviour. 
Since data gathered at the beginning of a restart tends to be noisy, 
we do not open the observation window immediately. 
To keep the feature vector of reasonable size, we use statistical 
measures of features (that is, the minimum over
the observation window, the maximum, the 
mean, the standard deviation and the last value
recorded). 
The list of features is shown in Table \ref{Table:featurevector}.
The only feature that takes more than constant time
to calculate is the \emph{log(WBE)} feature.
This is based on the Weighted Backtrack Estimator \cite{kil:sla:thi:wal}. 
This estimates search tree size using
the weighted sum: 
$\frac{\sum^{}_{d\in{D}}prob(d)(2^{d+1}-1)}{\sum^{}_{d\in{D}}prob(d)}$
where $prob(d)=2^{-d}$ and $D$ is the multiset of branches lengths visited. 
In \cite{hai:wal}, we extended WBE to support conflict driven backjumping.
As the new method requires $O(d)$ time and space, we only
compute it every $d$ conflicts.  
To deal with quick restarts, we wait until the observation window
fits within a single restart. In addition, we
exploit estimates from earlier restarts by augmenting the feature
vector with all the search cost predictions from previous
restarts.

\section {Experiments}
\label{Sec:results}

We ran experiments using MiniSat \cite{een:sor},
a state-of-the-art solver with clause
learning, an improved  version of VSIDS
and a geometrical restart scheme. 

We used a geometrical factor of 1.5, which is the default for MiniSat.
A geometrical factor of 1.2 gave similar results. 
We used three different ensembles of problems.
\begin{itemize}
\item \emph{rand:} 500 satisfiable
and 500 unsatisfiable random 3-SAT problems
with 200 to 550 variables and a clause-to-var ratio of 4.1 to 5.0. 
\item \emph{bmc:} 
250 satisfiable and 250 unsatisfiable 
software verification problems generated 
by CBMC\footnote{http://www.cs.cmu.edu/~modelcheck/cbmc/} for
on a binary search algorithm, using
different array sizes and number of loop unwindings. 
To generate satisfiable problems, faulty code 
that causes memory overflow was added. 
These problems create a very homogeneous ensemble.
\item \emph{fv:} 
56 satisfiable and 68 
unsatisfiable hardware verification problems
distributed by Miroslav 
Velev\footnote{http://www.miroslav-velev.com/sat\_benchmarks.html}. 
This is less homogeneous than the other ensembles.
\end{itemize} 
Since training examples can be scarce, we restricted
our training set to no more than 500 problems,
though we had far fewer for the hard verification problems. 
In the first part of our experiments, when restarts were
turned off, many of the hardware verification problems 
were not solved. Our results in this part will only compare 
the other datasets. When restarts were enabled, all three data sets were used.
In all experiments we used 10-fold cross validation, never using the same instance for both training and testing purposes.
We measured prediction quality by observing the percentage of 
predictions within a certain factor of the 
correct cost (the \emph{error factor}). 
For example, 80\% for error factor 2, denotes that 
for 80\% of the instances, the predicted search
cost was within a factor of 2 of the actual cost. 

\subsection{Search Without Restarts}

\begin{figure}
\centering
\subfigure[After 2000 backtracks]
{
    \label{Fig:lmp:wbe:rand:a}
    \includegraphics[width=4.00cm, angle=270]{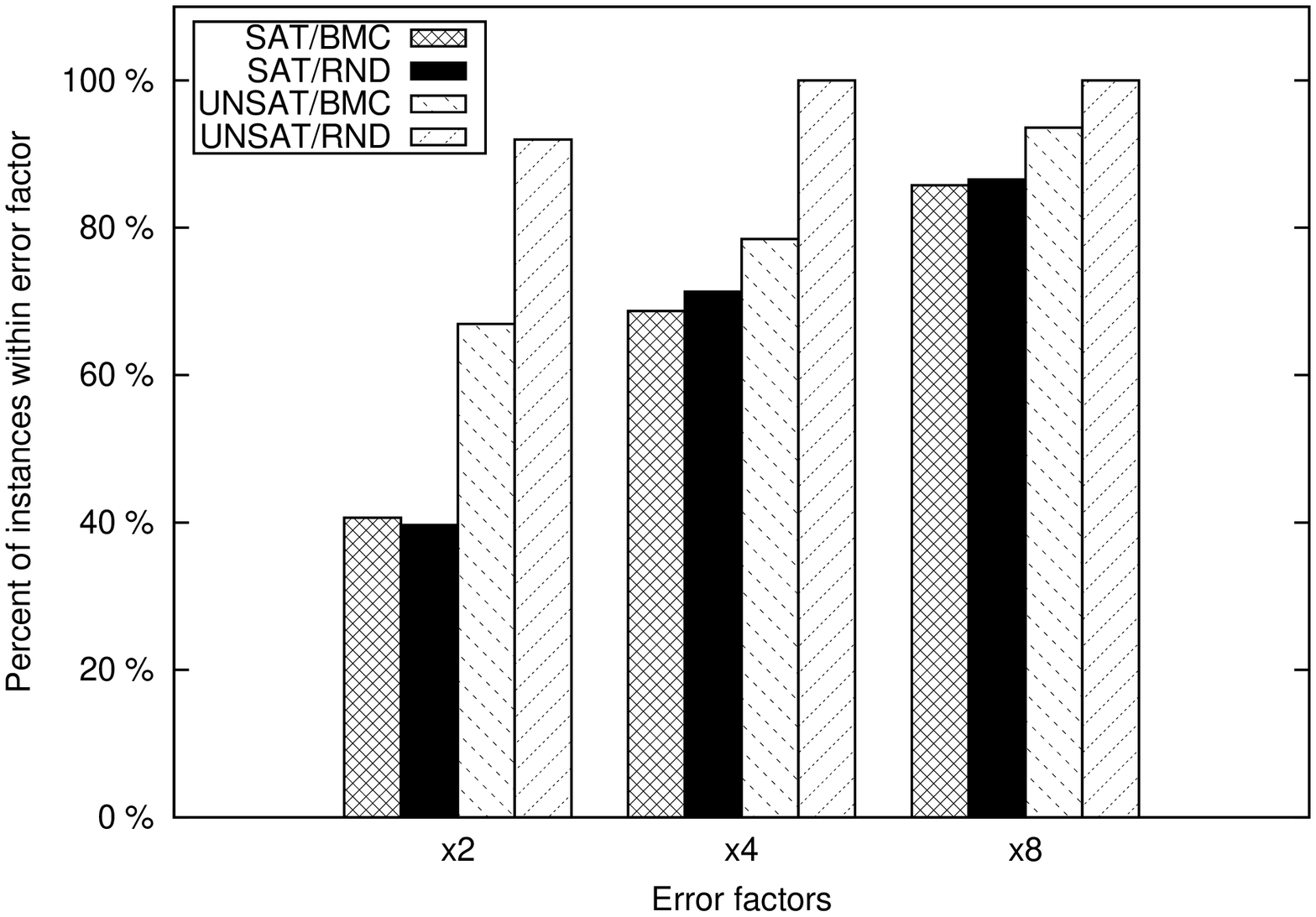}
}
\hspace{0.1cm}
\subfigure[After 35000 backtracks] 
{
    \label{Fig:lmp:wbe:rand:b}
    \includegraphics[width=4.00cm, angle=270]{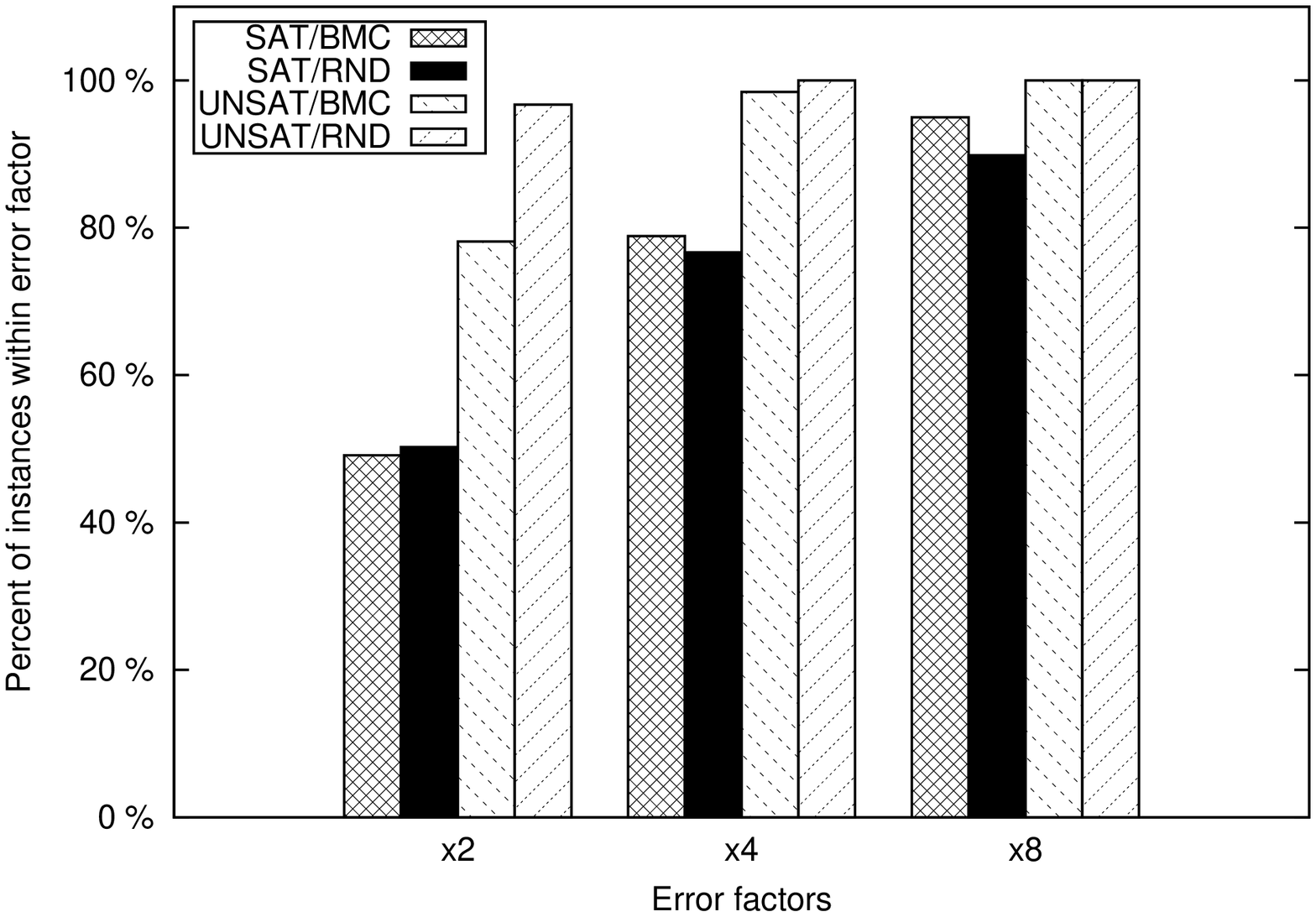}
}
\caption{Quality of prediction, without restarts, for the $rand$ and $bmc$ datasets.} 
\label{Fig:lmp:bmc:rand}
\end{figure} 

We queried our predictor at different points of the search, 
ranging from 2000 to 50000 backtracks. Comparisons 
of the performance of LMP for the \emph{rand} and 
\emph{bmc} data set are presented in Figure \ref{Fig:lmp:bmc:rand}.
Satisfiable problems are harder to predict for both $rand$ and $bmc$ datasets, 
due to the abrupt way in which search terminates with open nodes. 

\subsection{Search With Restarts}

\begin{figure}
\centering
\subfigure[$sat$] 
{
    \label{Fig:restarts:sat}
    \includegraphics[width=4.00cm, angle=270]{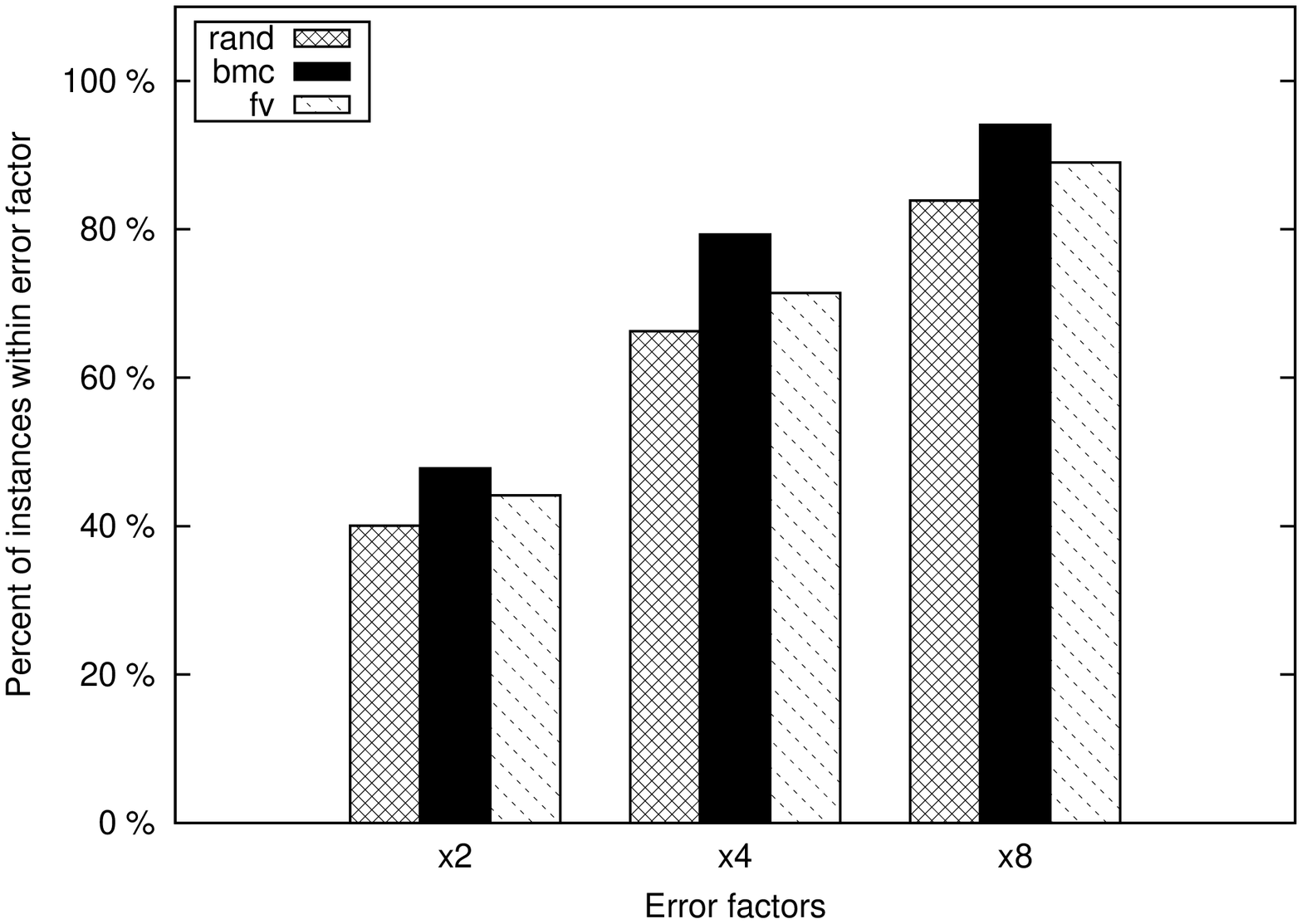}
}
\hspace{0.1cm}
\subfigure[$unsat$]
{
    \label{Fig:restarts:b}
    \includegraphics[width=4.00cm, angle=270]{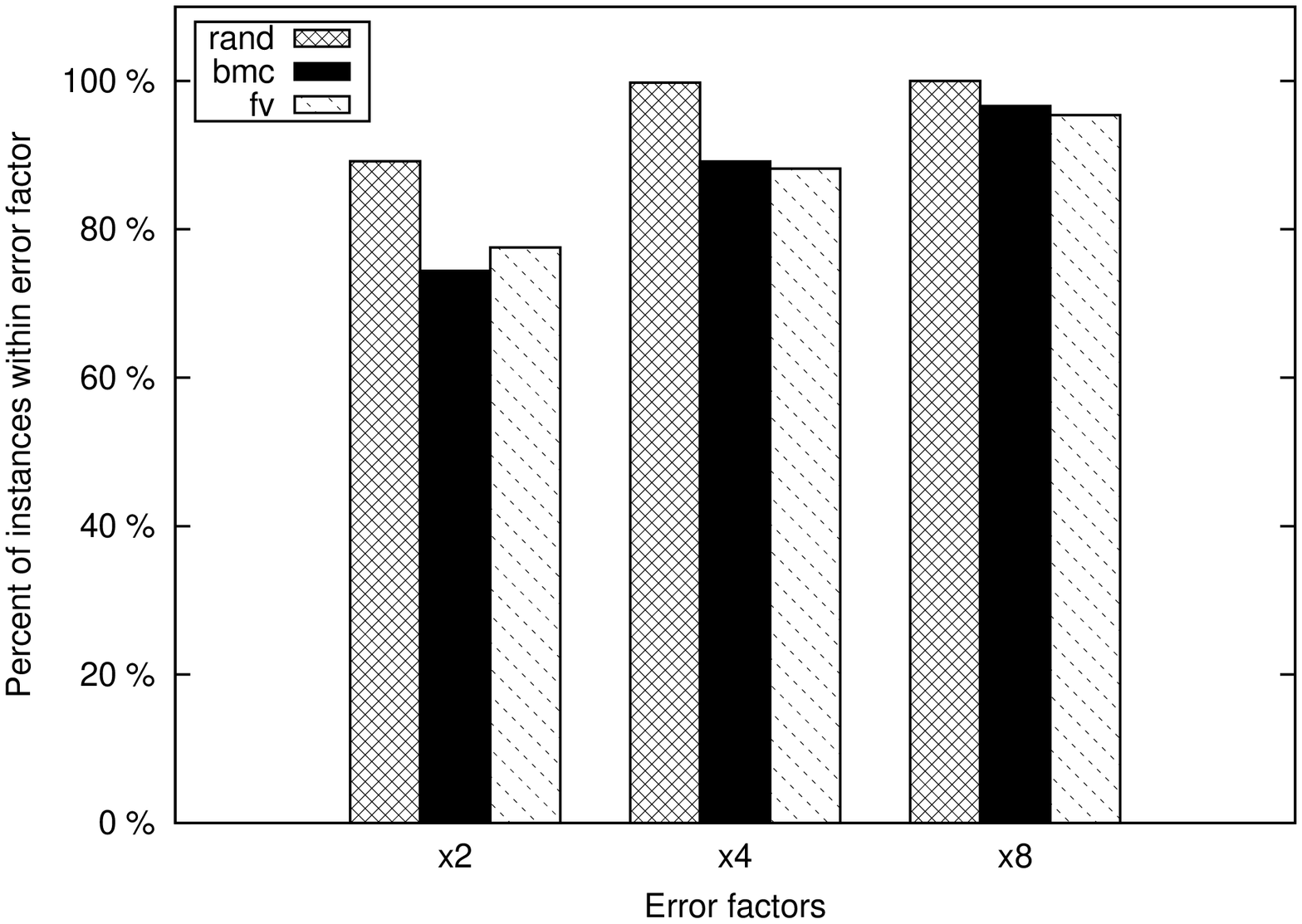}
}
\caption{Quality of prediction for the 3 different datasets when using restarts (after 2000 backtracks in the \emph{query restart})}
\label{Fig:restarts}
\end{figure}

With restarts, we have to use smaller observation windows
to give a prediction early in search as many early restarts are too small. 
Figure \ref{Fig:restarts} compares the quality of prediction of LMP 
for the 3 different datasets.
The quality of estimates improves with the $bmc$ data set
when restarts are enabled. We conjecture this is a result
of restarts before the observation 
window reducing noise.

\begin{figure}
\centering
\subfigure[$sat$] 
{
    \label{Fig:rolling:both:sat}
    \includegraphics[width=4.00cm, angle=270]{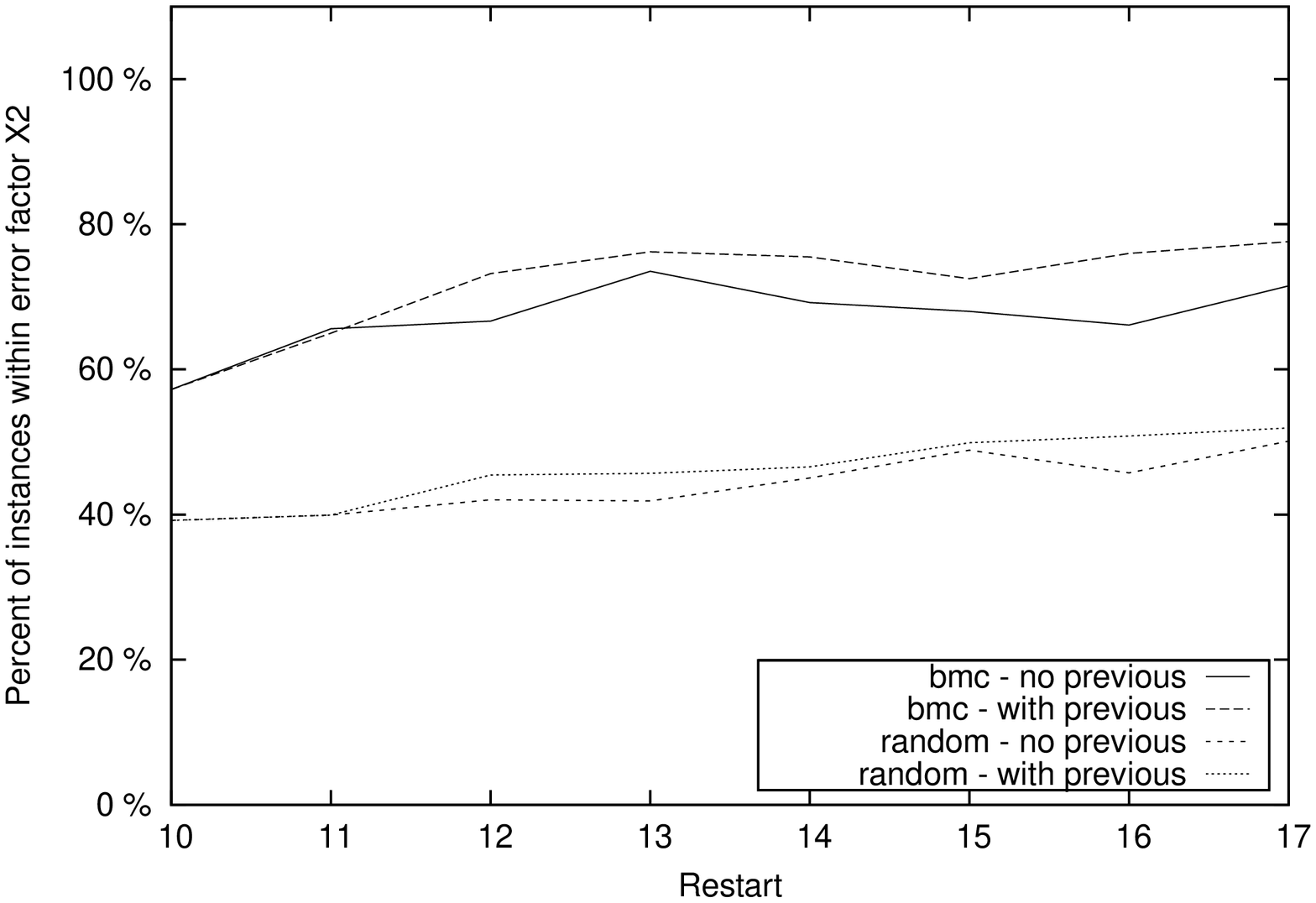}
}
\hspace{0.1cm}
\subfigure[$unsat$]
{
    \label{Fig:rolling:both:unsat}
    \includegraphics[width=4.00cm, angle=270]{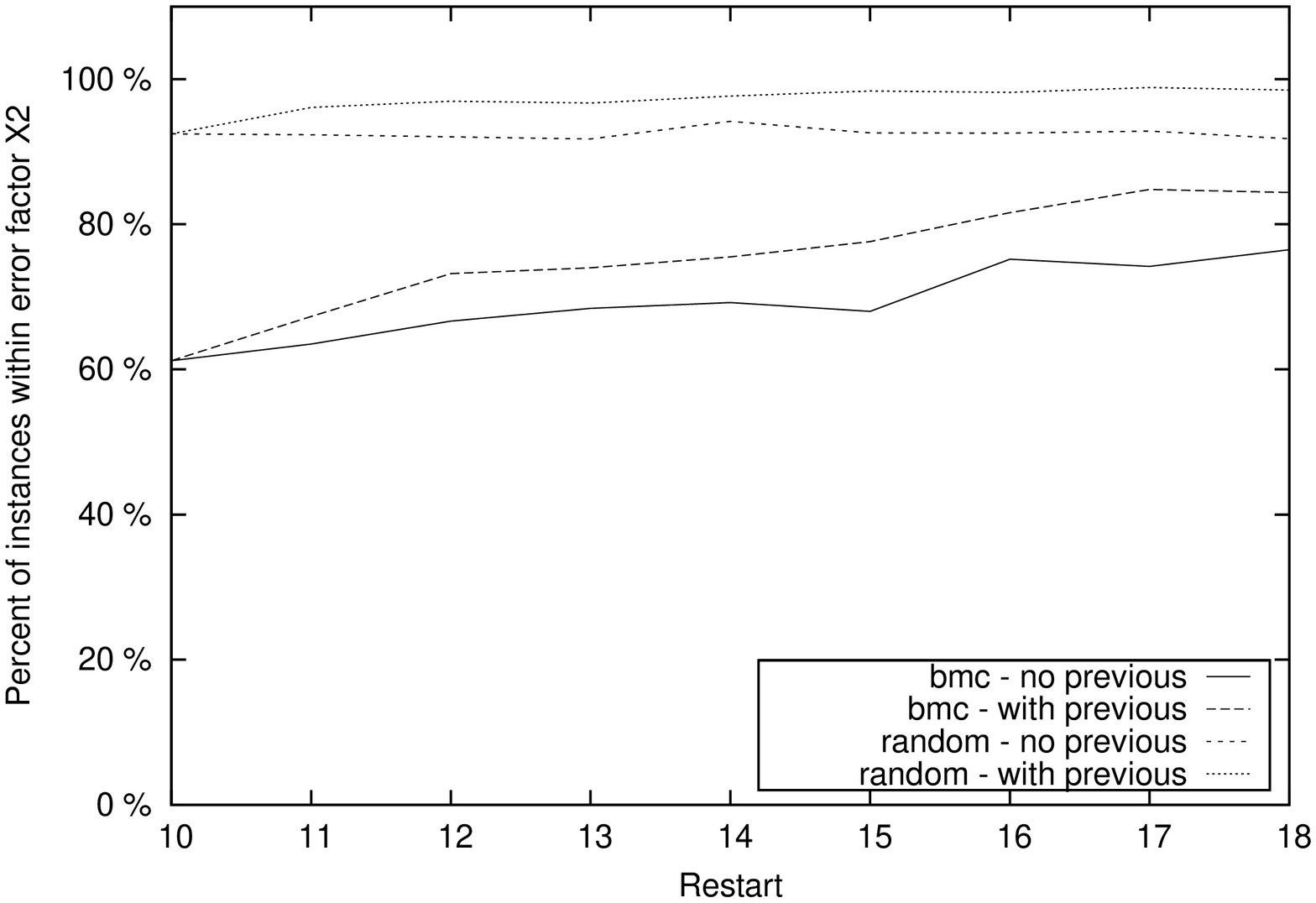}
}
\caption{The effect of using predictions from previous restarts. We compare the quality of prediction, through restarts, using two datasets ($bmc$,$rand$). The plots represent the percentage of instances within a factor of 2 from the correct size.  }
\label{Fig:rolling:both} 
\end{figure}

In order to see if predictions from previous restarts
improve the quality of prediction, 
we opened an observation window at every restart. The window 
size is $max(1000,0.01\cdot s)$ and starts after 
a waiting period of $max(500,0.02\cdot s)$, when $s$ is the 
size of the current restart. At the end of each observation 
window, two feature vectors were created. The first 
$\left(x_{r}\right)$ holds all features from Table \ref{Table:featurevector}, while the second $\left(\hat{x}_{r}\right)$ is defined as
$\hat{x}_{r}=\left\{x_{r}\right\}\cup\left\{f_{w_{1}}\left(x_{1}\right),f_{\hat{w}_{2}}\left(\hat{x}_{2}\right),\dots,f_{\hat{w}_{r-1}}\left(\hat{x}_{r-1}\right)\right\}$. Figure \ref{Fig:rolling:both} compares the two methods.
We see that predictions from earlier restarts improve the quality of later predictions but not greatly.

\subsection{Solver selection using LMP}

In our final experiment, we used these estimations
of search cost to improve solver performance. 
We used two different versions of MiniSat. Solver $A$
used the default MiniSat setting (geometrical factor of 1.5), 
while solver $B$ used a geometrical factor of 1.2. 
The challenge is to select which is faster at solving
a problem instance.

Table \ref{table:portfolio} describes the percentage improvement achieved by each of the following strategies. All values are fractions of the cost of solving the entire dataset, picking a solver randomly for each problem, with equal probability. Hence, for each dataset, $average(A,B)=1$:
\begin{itemize}
\item \emph{best:} Use an oracle to indicate which solver will solve the problem faster ($min(A,B)$).
\item \emph{LMP (oracle):} Use both solvers until each reaches the end of its observation window and generate a prediction, using two different models for $sat$ and $unsat$. Use a satisfiability oracle to indicate which model should be queried. Terminate the solver that is predicted to be worse. 
\item \emph{LMP (two models):} Use both solvers until each reaches the end of its observation window and generate a prediction, using two different models for $sat$ and $unsat$. Query both models and use the geometric mean as the prediction\footnote{We found this method to yield more accurate runtime estimations than using one model for both $sat$ and $unsat$ instances. For further details see \cite{hai:wal}.}. Terminate the solver that is predicted to be worse.
\end{itemize}

These results show that for satisfiable problems, where solver performance 
varies most, our method reduces the total cost. 
For unsatisfiable problems, where solver performance does 
not vary as much, our method does not improve search cost.
However, as performance does not change significantly on unsatisfiable instances, 
the overall impact of our method 
on satisfiable and unsatisfiable problems 
is positive. 
\begin{table}
\caption{Improvement in total search cost using different schemes}
\centering
\label{table:portfolio}
\begin{tabular}{ | l | l | c | c | c |}
\hline
\multicolumn{2}{| l |}{Dataset} & Best & LMP (oracle) & LMP(two models) \\
\hline
\hline
\multirow{2}{*}{$rand$} & sat 	& 0.591 & 0.930 & 0.895\\
												& unsat & 0.925 & 1.009 & 1.014\\\hline
\multirow{2}{*}{$fv$} 	& sat 	& 0.333 & 0.828 & 0.832\\
												& unsat & 0.852 & 1.006 & 1.033\\\hline
\multirow{2}{*}{$bmc$} 	& sat 	& 0.404 & 0.867 & 0.864\\
												& unsat & 0.828 & 0.997 & 1.004\\
\hline
\end{tabular}
\end{table}


\begin{thebibliography}{4} \itemsep=-1pt
%

\bibitem{alo:sie:sak}
Aloul, F., Sierawski, B., Sakallah, K.: 
Satometer: How much have we searched? In Design Automation Conf.,IEEE (2002) 737-742.

\bibitem{een:sor}
Een, N., Sorensson, N.: 
An extensible SAT-solver.
Theory and Applications of Satisfiability Testing, (2003), 502-518

\bibitem{hai:wal}
Haim, S., Walsh, T.:
SAT Solving Cost Estimation using Online Techniques, 
Technical Report 0805, UNSW, Australia, February 2008

\bibitem{hor:rua:gom:kau:sel:chi}
Horvitz, E., Ruan, Y., Gomes, C., Kautz, H., Selman, B., Chickering, M.: 
A Bayesian approach to tackling hard computational problems. 
Proc. the 17th Conf. on Uncertainty in Artificial Intelligence (UAI-2001), (2001)


\bibitem{kil:sla:thi:wal}
Kilby, P., Slaney, J., Thiébaux, S., Walsh, T.: 
Estimating Search Tree Size. 
Proc. of the 21st National Conf. of Artificial Intelligence, AAAI, (2006)

\bibitem{kok:shl}
Kokotov, D., Shlyakhter, I. Progress
bar for sat solvers. Unpublished manuscript,
http://sdg.lcs.mit.edu/satsolvers/progressbar.html. 2000.

\bibitem{ley:nud:sho}
Leyton-Brown, K., Nudelman, E., Shoham, Y.: 
Learning the Empirical Hardness of Optimization Problems: The Case of Combinatorial Auctions. 
Proc. of the 8th Int. Conf. on Principles and Practice of Constraint Programming, Springer-Verlag, (2002) 556-572


\bibitem{sam:mem}
Samulowitz, H., Memisevic, R.:
Learning to Solve QBF. 
In Proc. of 22nd Conf. on Artificial Intelligence (AAAI'07), (2007)

\bibitem{xu:hoo:ley}
Xu, L., Hoos, H.H., Leyton-Brown, K.: 
Hierarchical Hardness Models for SAT Principles and Practice of Constraint Programming, (2007), 696-711

\end{thebibliography}
\end{document}